\pdfoutput=1

\documentclass[11pt]{article}

\usepackage[]{EMNLP2024}

\usepackage{times}
\usepackage{latexsym}
\usepackage{amsmath}

\usepackage[utf8]{inputenc}

\usepackage{microtype}

\usepackage{inconsolata}

\usepackage[T1]{fontenc}
\usepackage{booktabs}
\usepackage{enumitem}

\newcommand{\strict}{\emph{Strict}\xspace}
\newcommand{\smallstr}{\emph{Strict-Small}\xspace}
\newcommand{\vision}{\emph{Multimodal}\xspace}
\newcommand{\multimodal}{\emph{Multimodal}\xspace}
\newcommand{\paper}{\emph{Paper}\xspace}

\def\best#1{\underline{#1}}
\def\bestall#1{\textbf{\best{#1}}}

\def\testsuite#1{\textsc{#1}}

\usepackage{todonotes}
\usepackage{array}
\usepackage{multirow,graphicx}
\usepackage{float}
\usepackage{cleveref}
\usepackage{xspace}
\usepackage{capt-of}

\title{Findings of the Second \includegraphics[height=0.4cm]{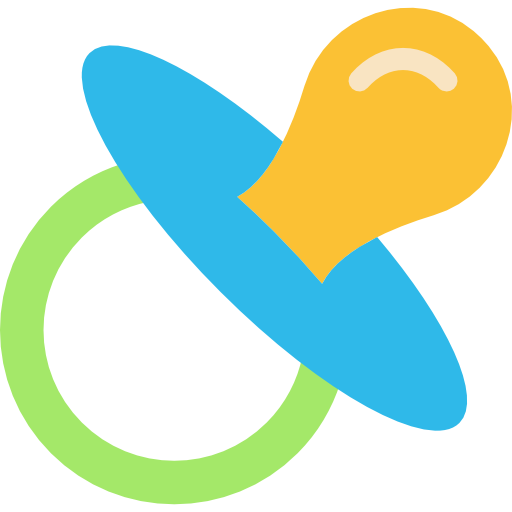} BabyLM Challenge:\\Sample-Efficient Pretraining on Developmentally Plausible Corpora}

\author{
         Michael Y. Hu$^{1}$ \ \ \ \ \ \ \ Aaron Mueller$^{2,3}$ \ \ \ \ \ \ \ Candace Ross$^4$ \\
         \textbf{Adina Williams}$^{4,7}$ \ \ \ \ \ \ \ \textbf{Tal Linzen}$^1$ \ \ \ \ \ \ \  \textbf{Chengxu Zhuang}$^6$ \ \ \ \ \ \ \ \textbf{Ryan Cotterell}$^8$\\
        \textbf{Leshem Choshen}$^{5,6}$\ \ \ \ \ \ \ \textbf{Alex Warstadt}$^{8}$\ \ \ \ \ \ \ \textbf{Ethan Gotlieb Wilcox}$^{9}$\\
        $^1$New York University\ \ \ \ \ \ \ $^2$Northeastern University\ \ \ \ \ \ \ $^3$Technion \ \ \ \ \ \ \
        $^4$Meta AI (FAIR) \\ $^5$IBM Research \ \ \ \ \ \ $^6$MIT \ \ \ \ \ $^7$ML Commons \\
        $^8$ETH Zürich \ \ \ \ \ $^9$Georgetown University\\
        \texttt{\href{mailto:michael.hu@nyu.edu }{michael.hu@nyu.edu}}
}

\begin{document}
\maketitle

\begin{abstract}
The BabyLM Challenge is a community effort to close the data-efficiency gap between human and computational language learners. Participants compete to optimize language model training on a fixed language data budget of 100 million words or less. This year, we released improved text corpora, as well as a vision-and-language corpus to facilitate research into cognitively plausible \emph{vision} language models. Submissions were compared on evaluation tasks targeting grammatical ability, (visual) question answering, pragmatic abilities, and grounding, among other abilities. Participants could submit to a 10M-word text-only track, a 100M-word text-only track, and/or a 100M-word and image multimodal track. From 31 submissions employing diverse methods, a hybrid causal-masked language model architecture outperformed other approaches. No submissions outperformed the baselines in the multimodal track. In follow-up analyses, we found a strong relationship between training FLOPs and average performance across tasks, and that the best-performing submissions proposed changes to the training data, training objective, and model architecture. This year's BabyLM Challenge shows that there is still significant room for innovation in this setting, in particular for image-text modeling, but community-driven research can yield actionable insights about effective strategies for small-scale language modeling.

\end{abstract}

\section{Introduction}

This paper describes the second BabyLM Challenge and its findings. 
The broader goals and motivation of the challenge have remained constant since the first iteration last year. At the heart of both this year's and last year's challenge is the observation that children are incredibly data-efficient language learners, whereas artificial neural-network-based language models are not. On the one hand, children are exposed to less than 100 million word tokens by the age of 13 \citep{gilkerson2017mapping}, at which point they have mastered their native language(s). On the other hand, today's ANN-based language models are trained on trillions of words---five to six orders of magnitude more than the typical human language learner. For a more in-depth discussion on the issue of data efficiency, see the findings of last year's challenge \citep{warstadt2023findings} as well as \citet{wilcox2024bigger}, a position piece written by many of the challenge organizers.

The learning discrepancy between humans and models raises two important questions: First, how is it that humans are able to learn language so efficiently? And second, what insights from human language learning can be used to improve language models? It is our hope that by creating a platform for interested parties to experiment with data-limited and cognitively inspired language modeling, we can continue to make progress on these interrelated questions. In particular, our goal with BabyLM is to contribute to:
\begin{enumerate}
    \item Building more cognitively and developmentally plausible models of human language acquisition and processing, which can be used for the scientific study of language.
    \item Optimizing training pipelines prior to scaling, allowing for faster iteration on architectures and hyperparameters.
    \item Enabling research on language model training to a wider group of interested researchers, beyond highly-funded industry labs.
\end{enumerate}

The main difference between this year's and last year's challenge is twofold: First, this year we allowed participants to bring their own datasets, as long as they stayed within the 100 million word limit for our \strict track, or the 10 million word limit for our \smallstr track. The motivation behind this decision is that pretraining data quality has been linked to large improvement gains in at-scale language models \citep{gunasekar2023textbooksneed}, so this year we allowed participants to improve the training data beyond the provided dataset, which was effectively a dataset baseline. Second, this year included a \vision{} track, in which participants trained on aligned text-image data, and tested their models in a novel text-image evaluation pipeline. Non-linguistic information, such as visual input, potentially plays a large role in child language acquisition. While visual input is not inherently necessary for successful language acquisition (for example, blind children learn language largely without issue), visual grounding has been linked to faster language learning \cite{perez1992pragmatic, campbell2024role}. Furthermore, visual grounding has long been hypothesized to aid word learning: children learn nouns more easily than verbs \cite{gentner1982nouns, mcdonough2011image}, arguably because the former are more easily linked to visual stimuli than the latter. Additionally, children learn concrete nouns easier than abstract nouns \cite{bergelson2013acquisition}. However, visual grounding also presents several challenges: Words may be time-delayed with respect to their referents, or one word may be uttered in a context with multiple competing possible referents. With this in mind, our hope was that the \vision track would help to explore the space of possible computational models for visual grounding during language acquisition.


\paragraph{Findings and takeaways.}
This year, we received 31 submissions from 17 different countries making diverse contributions. Examples included submissions proposing novel architectures, new training objectives, innovating on knowledge distillation methods, and proposing curriculum learning methods, among others. We conduct a meta-analysis of the results, which yields several concrete recommendations. The best-performing submissions constructed their own training datasets, proposed new model architecture, or new training objectives. Performance on the BabyLM evaluations also correlated strongly with training FLOPs, suggesting that high-compute training regimes still tend to reliably perform better, even in low-data settings. The BabyLM research community also showed growing attention to tokenization and multilingual language modeling, while maintaining interest in curriculum learning and applying linguistic biases to language models.

Our data (pretraining corpora and evaluation data; \href{https://osf.io/ad7qg/}{[link]}), preprocessing code \href{https://github.com/babylm/babylm_data_preprocessing}{[link]}, baselines \href{https://huggingface.co/babylm}{[link]} and evaluation pipeline \href{https://github.com/babylm/evaluation-pipeline-2024}{[link]} are all publicly available. We also release the submitted models of those who agreed to release them, along with their hyperparameters and results \href{https://docs.google.com/spreadsheets/d/182IjCUiaVYSuJq9GAwZeeb-50bxBlY4qEMOdiCh6i-g/edit?gid=0#gid=0}{[link]}. The leaderboard may be found here \href{https://huggingface.co/spaces/babylm/leaderboard-2024}{[link]}.


\section{Competition Details} \label{sec:instructions}

\paragraph{Tracks.}

The second BabyLM Challenge included three competition tracks: \textbf{\strict}, \textbf{\smallstr}, and \textbf{\vision}. Additionally, we opened a standalone \textbf{\paper} track, accepting research related to cognitive modeling with language models or small-scale pretraining, similar to a workshop. 

The \strict{} and \smallstr{} tracks required that submissions be trained on 100M words or less and 10M words or less, respectively. These tracks no longer required that participants use the fixed dataset from last year's challenge, although we still provided an updated version of this dataset, described in Section \ref{sec:data}. Models in these tracks were evaluated on language-only evaluation tasks.

In the \vision{} track, participants trained multimodal image-text models. Participants were allowed to use any model and training procedure they desired, as long as the model could assign (pseudo) log-likelihoods to strings of text, conditioned on an image. Again, participants were free to construct their own datasets, including unlimited visual inputs, as long as the text data was within a 100M word budget. To facilitate easier participation in this track, we released a suggested multimodal dataset that consisted of 50\% text-only and 50\% paired image-text data. Submissions to this track were evaluated on both language-only and additional multimodal tasks.


\section{Pretraining Corpus}\label{sec:data}

\begin{table*}[t]
    \centering
    \resizebox{\linewidth}{!}{
    \begin{tabular}{llrrr}
    \toprule
    Dataset & Description & \# Words (multimodal) & \# Words (strict) & \# Images\\
    \midrule
    Localized Narratives\textsuperscript{a} & Image Caption & 27M & -- & 0.6M \\
    Conceptual Captions 3M\textsuperscript{b} & Image Caption & 23M & -- & 2.3M \\
    CHILDES\textsuperscript{c} & Child-directed speech & 14.5M & 29M & -- \\
    British National Corpus (BNC), dialogue portion\textsuperscript{d} & Dialogue & 4M & 8M & -- \\
    Project Gutenberg (children's stories)\textsuperscript{e} & Written English & 13M & 26M & -- \\
    OpenSubtitles\textsuperscript{f} & Movie subtitles & 10M & 20M & -- \\
    Simple English Wikipedia\textsuperscript{g} & Written Simple English & 7.5M & 15M & -- \\
    Switchboard Dialog Act Corpus\textsuperscript{h} & Dialogue & 0.5M & 1M & -- \\
    \midrule
    \emph{Total} & -- & 100M & 100M & 2.9M\\
    \bottomrule
    \end{tabular}}
    \caption{Datasets for the multimodal and strict tracks of the 2nd BabyLM Challenge. Word counts are approximate and subject to slight changes. 
    \textsuperscript{a}\citet{LocalizedNarratives} \ \ \ 
    \textsuperscript{b}\citet{CC3M} \ \ \
    \textsuperscript{c}\citet{macwhinney2000childes} \ \ \
    \textsuperscript{d}\citet{bncconsortium2007british} \ \ \
    \textsuperscript{e}\citet{gerlach-2018-gutenberg} \ \ \
    \textsuperscript{f}\citet{lison-tiedemann-2016-opensubtitles2016} \ \ \
    \textsuperscript{g}\href{https://dumps.wikimedia.org/simplewiki/}{\url{https://dumps.wikimedia.org/simplewiki/}} \ \ \
    \textsuperscript{h}\citet{Stolcke-etal:2000} \ \ \
    \looseness=-1}
    \label{tab:data}
\end{table*}

This year, we updated the text-only dataset from the previous competition and provided a novel image-text dataset for the \vision{} track. Data for both the text-only and multimodal datasets can be downloaded from \url{https://osf.io/ad7qg/}.

For the text-only dataset updates, we increased the proportion of child-oriented data (counting both transcribed speech and written data) to 70\% up from 39\% last year, and we increased transcribed speech data to 58\% up from 55\% last year. We have eliminated the Wikipedia portion of the data (except for Simple English Wikipedia) due to being the only non-spoken and non-child-level data, and we have eliminated the QED portion due to quality issues. We have also reduced our reliance on OpenSubtitles, which can include scripted speech, which is arguably less ecologically valid than other spoken sources.
CHILDES now comprises a significantly larger portion of the new dataset. We use the entire available English portion of CHILDES including both caregiver and child utterances, increasing the proportion of child-oriented discourse from 5\% last year to 29\%.\footnote{We thank Brian MacWhinney (personal correspondence) for alerting us to the existence of this additional CHILDES data.}
We also replaced last year's children's stories and Project Gutenberg data with a custom children's stories dataset sourced entirely from Project Gutenberg. We select child-appropriate books using the provided \texttt{subject} metadata, and then select the 1000 most common books, giving us a combined corpus of 26M words. For more details about other data sources, see \citep{warstadt2023findings}.

In addition, we provide a novel image-text dataset to facilitate easier participation in the \vision track. This dataset has two components: First, we provide 50M words of text-only data, drawn from the 100M BabyLM corpus via stratified sampling (that is, we preserve the relative distribution from the different data sources). Second, we provide paired text-image data that includes 50M words of text. This paired data comes from two sources: 27M words from the Localized Narratives dataset \citep{PontTuset_eccv2020} and 23M words from the Conceptual Captions 3M (CC3M) dataset \citep{sharma2018conceptual}.
For the Localized Narratives dataset, we used the text captions and the images from the MS-COCO~\cite{lin2014microsoft} and Open Images~\cite{OpenImages} subsets.
For the CC3M dataset, we used the image-caption pairs whose images were still valid in January 2024.
In the OSF directory at the above link, we provided scripts to download the images.
Table \ref{tab:data} gives an overview of the datasets comprising the BabyLM pretraining set, and descriptions of each data source are provided in \Cref{app:data_sources}.


\subsection{Preprocessing}

We released train, validation, and test splits for each of the ten data sources in \strict and \smallstr in proportions 83.3\%/8.3\%/8.3\%, respectively. The 10M word \smallstr training set is sampled randomly from the \strict training set: after preprocessing, we downsampled and split each source by randomly sampling chunks of 2000 lines or longer. 
The code and instructions for downloading and preprocessing the raw data are publicly available.\footnote{\url{https://github.com/babylm/babylm_data_preprocessing}}

We performed minimal preprocessing in terms of filtering and reformatting text. 
Notably, we preserved newlines, meaning newlines do not consistently delimit documents, paragraphs, or sentences, as in some pretraining datasets. 
We used WikiExtractor \citep{Wikiextractor2015} to extract text from the \texttt{xml} Simple English Wikipedia dump dated 2022-12-01. 
We removed \texttt{<doc>} tags in Simple English Wikipedia and selected the spoken subset of the BNC by taking only lines from the \texttt{xml} containing the \texttt{<stext>} tag and extracting the text from the \texttt{xml}. 
We used code by \citet{gerlach2020standardized} to download and preprocess data from Project Gutenberg, which we additionally filtered to contain only English texts by authors born after 1850. 
The OpenSubtitles and Wikipedia portions of the pretraining corpus were shared with us in preprocessed form, having had duplicate documents removed from OpenSubtitles and preprocessing steps performed to Wikipedia similar to our Simple English Wikipedia procedure.\footnote{We thank Haau-Sing Li for allowing us to use this preprocessed data.} 
We used regular expressions to remove speaker and dialog act annotations from the Switchboard Dialog Act Corpus and annotations from the CHILDES data.
We preserved speaker annotations and scene descriptions from CHILDES.
We performed no preprocessing on the remaining datasets.

\section{Evaluation and Submission}

As in last year, we distributed a shared evaluation pipeline based on the LM Evaluation Harness \citep{eval-harness}. For the \strict{} and \smallstr{} tracks, evaluation tasks were largely the same as the previous year: we used BLiMP \citep{warstadt-etal-2020-blimp-benchmark-}, the BLiMP Supplement \citep{warstadt2023findings}, and a subset of (Super)GLUE tasks \citep{wang-etal-2019-superglue,wang-etal-2018-glue-} as the public evaluation set. BLiMP measures whether LMs prefer grammatical to minimally-differing ungrammatical sentences (i.e., minimal pairs) and spans a range of grammatical phenomena including subject-verb agreement, binding, and control/raising constructions. The BLiMP supplement is a disjoint subset of minimal pairs designed specifically for last year's BabyLM Challenge to test linguistic knowledge not covered by BLiMP, such as dialogue and pragmatics. (Super)GLUE is designed to measure natural language understanding across a diverse array of subtasks; its tasks include question answering and natural language inference, among others.

For the \vision{} track, participants were required to evaluate on the evaluation benchmarks from the text tracks; this was to establish whether training on image data facilitated sample-efficient language modeling. In addition, we included a suite of multimodal evaluation tasks. The public evaluation datasets included Visual Question Answering (VQA; \citealp{Antol_2015_ICCV,Goyal_2017_CVPR}) and Winoground \citep{Thrush_2022_CVPR}. VQA measures whether vision-language models (VLMs) prefer correct answers to questions about visual inputs, and Winoground measures whether LMs prefer accurate descriptions of images among minimally differing options (e.g., given an image of dirt on top of a light bulb, does the VLM prefer ``a lightbulb on top of dirt'', or ``dirt on top of a lightbulb'', and vice versa given another image where the lightbulb is on top of dirt).

This year, we used the Elements of World Knowledge (EWoK) dataset \citep{ivanova2024elements} as the hidden task for the text tracks. This task measures pragmatic, commonsense, and discourse knowledge. For the \vision{} track, the hidden task was DevBench \citep{tan2024devbench}; this benchmark contains subtasks targeted at evaluating visual and linguistic abilities that emerge at different stages of children's development, including subtasks where (i) the model must pick the correct image associated with a given word; (ii) the model must pick the correct image corresponding to a sentence; and (iii) the model must assign appropriately higher or lower similarity scores to more or less similar images. The data for these tasks was released two weeks before the model submission deadline. We selected these tasks based on whether they capture distinct phenomena from the public evaluation tasks, such that optimizing only for individual tasks or narrow subsets of linguistic competencies would not be overly rewarded.

Most of the evaluation tasks were zero-shot. Zero-shot evaluation entails comparing the probabilities of different sequences of text. Thus, all submitted models were required to assign a (pseudo) log-likelihood to a sequence of tokens. Additionally, the (Super)GLUE tasks required fine-tuning a classification head appended to the model. Models did not need to \emph{generate} sequences for any evaluation task; thus, both autoregressive and masked language modeling architectures could be used. 

\subsection{Evaluation Pipeline}\label{sec:eval-pipeline}
We provided code to unify the evaluation setup across submissions. This was released as a public repository on GitHub.\footnote{\url{https://github.com/babylm/evaluation-pipeline-2024}} The evaluation pipeline supports models implemented in HuggingFace, including Transformer-based architectures, structured state space models (e.g., Mamba; \citealp{gu2024mamba}), and recurrent neural networks \cite{peng-etal-2023-rwkv}, among other architectures. Note, however, that we did not restrict the model submissions to HuggingFace-based models; participants were allowed to use their own evaluation setup if desired, so long as they were able to produce predictions in the expected format.\footnote{Upon release of the evaluation pipeline, we announced that we would provide support as needed to teams training LMs not based in HuggingFace.} For model and result submissions, users were required to (i) upload a link to their model (on any file-hosting service), and (ii) provide model predictions for each example of each task; we provided a template specifying the format of the predictions file in the evaluation pipeline repository.

\paragraph{Data preprocessing.} 
NLP tasks in our evaluation pipeline often contained vocabulary that is not contained in the BabyLM pretraining corpora. To address this mismatch, we filtered each evaluation task according to its lexical content. We first computed two vocabularies by collecting all words that appear at least twice in the \smallstr{} corpus and collecting all words that appear at least twice in the \vision{} corpus. Then, we took the intersection of these two vocabularies to obtain the final vocabulary. Finally, we iterated through each example in each evaluation task; if an example contained any words that appeared less than twice in the final vocabulary, we filtered the example. Otherwise, each dataset is presented in its original format. See Table~\ref{tab:eval-data-size} in Appendix~\ref{app:eval-data-size} for details on the size of the filtered datasets.

\subsubsection{Evaluation Paradigms}
\paragraph{Zero-shot evaluation.} For zero-shot tasks---all of them except (Super)GLUE---we modified the \texttt{lm-eval-harness} repository, originally by EleutherAI \citep{eval-harness}. This provides functionality for scoring autoregressive decoder-only LMs and encoder-decoder LMs. For encoder-only LMs, we modified the repository to support masked language model scoring as described in \citet{salazar-etal-2020-masked}, and as updated by \citet{kauf-ivanova-2023-better}.\footnote{We used the implementation of \citet{misra2022minicons} in the \texttt{minicons} library.} We also modified the pipeline to support multimodal models and tasks.

\paragraph{Finetuning.} Prior to the challenge, we experimented with zero-shot learning and few-shot in-context learning for (Super)GLUE. However, this often resulted in random-chance accuracies from our baselines; we therefore employed finetuning. While finetuning technically adds to the training set size, we consider this acceptable, as finetuning on a single GLUE or MSGS task does not meaningfully add to the domain-general linguistic abilities of language models. For tasks requiring finetuning---namely, (Super)GLUE \citep{wang-etal-2018-glue,wang-etal-2019-superglue}---we base our scripts on HuggingFace's example finetuning scripts for text classification.\footnote{\url{https://github.com/huggingface/transformers/blob/211f93aab95d1c683494e61c3cf8ff10e1f5d6b7/examples/pytorch/text-classification/run_glue.py}} We modified the script from last year's pipeline to work with more recent versions of HuggingFace transformers. We provided a default set of hyperparameters that we found to work well across our baseline models, though participants were allowed to modify hyperparameters if they wished. We also provided support for fine-tuning models via low-rank adapters (LoRA; \citealp{hu2022lora}). This enabled the possibility of faster and more compute-efficient model adaptation for our tasks.

\subsection{Submission process}
\paragraph{Submission format.} The submission form was hosted via OpenReview. We required a link to the models, as well as a link to the predictions of these models for all examples for all tasks. The predictions file was formatted as a JSON; each example had an entry with an example ID as its key, and the the prediction of the model as its value. For classification tasks, a prediction was a label ID integer. For zero-shot tasks, predictions were the string that received the highest probability according to the model. The submission process for the competition consisted of three components, which are outlined below:

\paragraph{Paper submission.} Each participant submitted a paper detailing their research, methodology, experimental design, and key findings. This was required for all participants, even if they did not submit a model to compete in the challenge.

\paragraph{Artifact submission.} In addition to the paper, participants who opted to compete and adhere to the competition rules were required to provide supplementary materials, including model outputs, checkpoints, and pretraining data (unless the default pretraining dataset was used). Participants were also required to upload their predictions for all evaluation tasks. 

\paragraph{Submission form.} To facilitate comparability and reproducibility, participants were asked to fill in a standardized form that captured model metadata, including hyperparameters, submission descriptions, and links to custom data if the standard corpus was not used.

\subsection{Baselines} \label{sec:baselines}
As opposed to last year's baselines, which were selected and trained relatively naively, this year's baselines were based on the architectures of winning submissions from last year's challenge. For the \strict{} and \smallstr{} tracks, we released the following baselines: LTG-BERT (encoder-only; \citealp{samuel-etal-2023-trained}) and BabyLlama (decoder-only; \citealp{babylm2023babyllama}). Although a variant of LTG-BERT (called ELC-BERT) won last year's challenge \citep{charpentier_samuel_2023_layers}, \citet{wilcox2024bigger} showed that similar performance on BabyLM evaluations can be achieved without the additional modifications of ELC-BERT. Thus, we chose LTG-BERT as the baseline, as it is a simpler model. BabyLlama is architecturally similar to Llama (albeit with far fewer parameters), and is additionally trained using knowledge distillation. For the \vision{} track, we released vision language models based on GIT~\cite{wang2022git} and Flamingo~\cite{alayrac2022flamingo} architectures, both of which are autoregressive.

\paragraph{Implementation details.} For LTG-BERT, we initially used the code provided in the repository linked in \citet{samuel-etal-2023-trained}, but we encountered unstable training due to loss spikes with this setup. 
We therefore used the LTG-BERT model released on HuggingFace, and trained it using the HuggingFace trainer. While training was still relatively unstable compared to other architectures, this procedure yielded performance in the expected range relative to other baselines. For BabyLlama, we use the code from the repository linked in \citet{babylm2023babyllama}, with small changes for compatibility with this year's BabyLM corpus. For the GIT and Flamingo baselines, we adapt the implementation of \citet{zhuang-etal-2024-visual}. Note that these baselines are not necessarily meant to achieve high scores on our evaluation tasks; rather, they are meant to encourage participants to innovate and improve beyond naive applications of existing methods.

\begin{figure}[t]
    \centering
    \includegraphics[width=\linewidth]{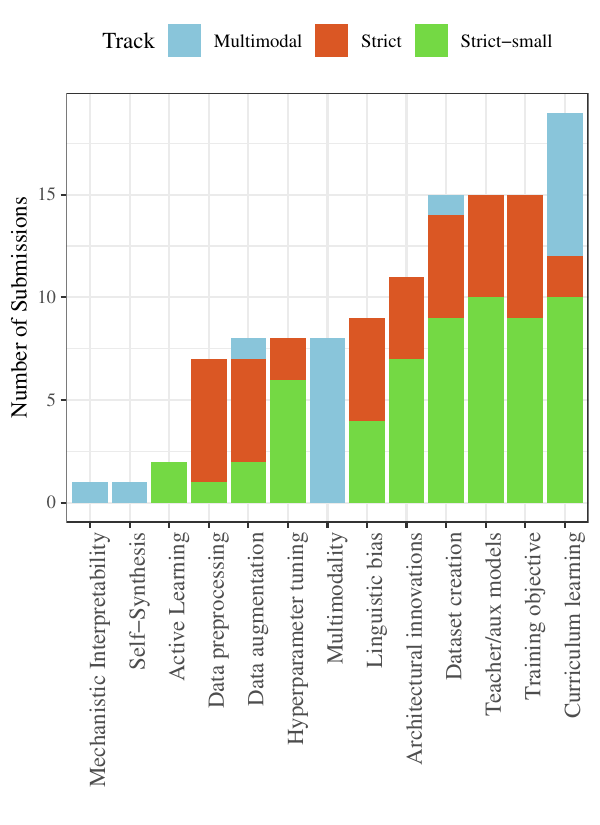}
    \caption{A breakdown of the various approaches used in the 2024 BabyLM challenge, organized by category and track. Curriculum learning again takes the top spot as the most popular approach, followed by training objective innovations.}
    \label{fig:submissions_by_approach}
\end{figure}

\section{Competition Results} \label{sec:results}

\begin{table}[]
    \centering
    \begin{tabular}{lcc}
    \toprule
        & \bf \# Models & \bf \# Participants\\\midrule
        \vision & 8 & 3 \\
        \smallstr & 35 & 18 \\
        \strict & 19 & 11 \\
        \midrule
        \it Total & \it 64 & \it 31 \\\bottomrule
    \end{tabular}
    \caption{Total number of models and participants per track. Participants who submitted to multiple tracks are counted once in the total. Two models were submitted to the \paper{} track only.}
    \label{tab:submission_counts}
\end{table}

In this section, we discuss the overall results of the competition (\S \ref{subsec:results}), track winners (\S \ref{sec:winning-submissions}), and this year's Outstanding Papers (\S \ref{subsec:outstanding}). 

We received 31 papers and 64 models in total, with two models submitted to the paper track. Table \ref{tab:submission_counts} shows the submission counts for each track. Despite efforts to make text--vision pretraining as accessible as possible, only three teams submitted to the \multimodal{} track, for a total of 8 model submissions. As none of these submissions outperformed our baselines, we decided not to award a winner in this track. Despite this disappointment, we hope that our datasets and evaluation resources serve as a basis for further exploration of text-image models in the years to come.

We found that many submissions focused their efforts on similar techniques. To better quantify this, we devised, in Figure \ref{fig:submissions_by_approach}, a typology of the most common approaches and assigned each submitted model one or more labels. \S\ref{subsec:common_methods} provides more detailed descriptions of each approach, as well as results indicating which ones were most effective.

All participants are affiliated with universities or independent research institutions. Participants' home institutions are located in 16 different countries. The number of participants by country is as follows (multinational submissions are counted more than once): Germany (8), United States (6), Netherlands (4), Italy (2), UK (2), Canada (1), China (1), Greece (1), Hungary (1), Iran (1), Israel (1), Japan (1), Norway (1), Singapore (1), Sweden (1), Switzerland (1), and Taiwan (1).

\begin{figure*}[t]
    \centering
    \includegraphics[width=0.29\linewidth]{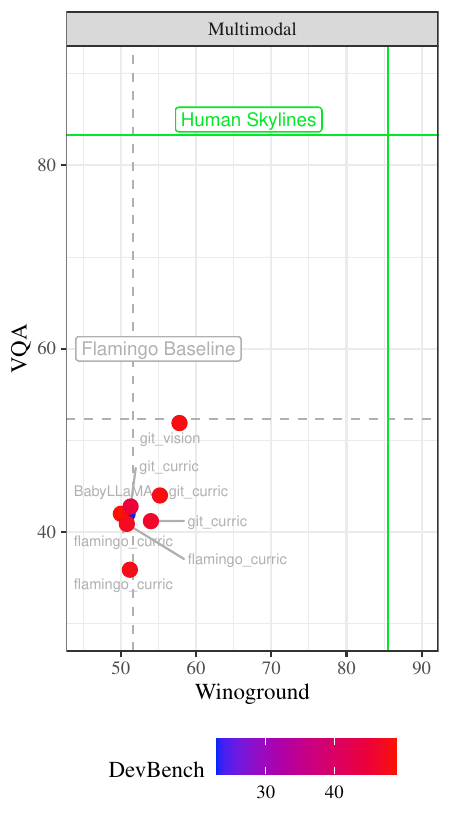}
    \includegraphics[width=0.68\linewidth]{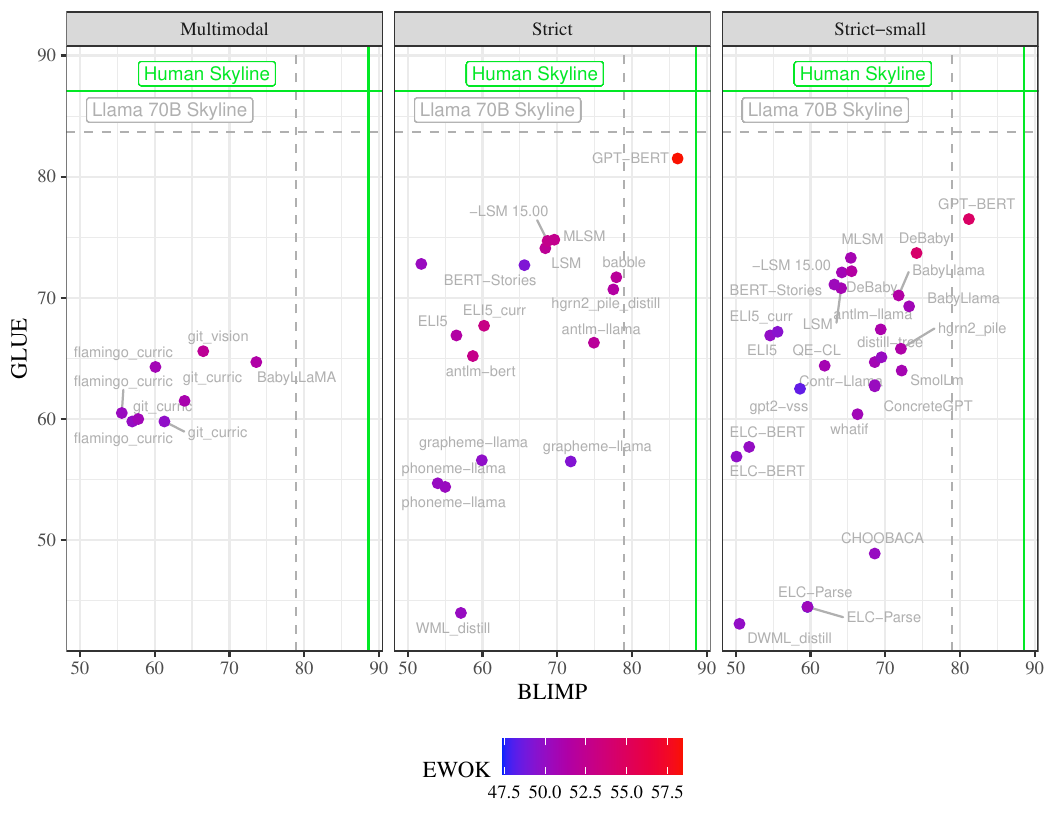}
    \caption{Overall results: At left, multimodal models on multimodal tasks; at right, all models on text tasks. N.B. Human scores for multimodal evals differ somewhat from how we evaluate our models. 
    }
    \label{fig:overall-results}
\end{figure*}

\subsection{Overall Results \& Track Winners} \label{subsec:results}

\begin{table*}[ht]
\def\supptask#1{\rotatebox[origin=c]{90}{\parbox[t]{6em}{\small#1}}}
{\centering
\begin{tabular}{lllllllllll}
\toprule
    \textbf{} & \textbf{Model} & 
    \supptask{BLiMP} & 
    \supptask{BLiMP\\Supplement} & 
    \supptask{(Super)GLUE} & 
    \supptask{EWoK} & 
    \supptask{\emph{Text\\Average}} &
    \supptask{VQA} & 
    \supptask{Winoground} &
    \supptask{DevBench} &
    \supptask{\emph{Vision\\Average}} \\ 
\midrule
    \parbox[t]{2mm}{\multirow{4}{*}{\rotatebox[origin=c]{90}{\small{Strict}}}} 
    & GPT-BERT &  \bestall{86.1} & \bestall{76.8} & \bestall{81.5} & \bestall{58.4} & \bestall{75.7} & -- & -- & -- & -- \\ 
    ~ & BabbleGPT & 77.9 & 69.5 & 71.7 & 52.0 & 67.8 \\ 
    ~ & MLSM & 69.6 & 65.4 & 74.8 & 52.6 & 65.6 & -- & -- & -- & -- \\ 
    ~ & \emph{Best baseline: LTG-BERT} & 69.2 & 66.5 & 68.4 & 51.9 & 64.8 & -- & -- & -- & -- \\ 
\midrule
    \parbox[t]{2mm}{\multirow{4}{*}{\rotatebox[origin=c]{90}{\small{Strict-small}}}} 
    & GPT-BERT & \best{81.2} & \best{69.4} & \best{76.5} & \best{54.6} & \best{70.4} & -- & -- & -- & -- \\ 
    ~ & DeBaby & 74.2 & 63.7 & 73.7 & 54.3 & 66.5 & -- & -- & -- & --\\ 
    ~ & BabyLlama-2 & 71.8 & 63.4 & 70.2 & 51.5 & 64.2 & -- & -- & -- & -- \\ 
    ~ & \emph{Best baseline: BabyLlama} & 69.8 & 59.5 & 63.3 & 50.7 & 61.6 & -- & -- & -- & -- \\ 
\midrule
    \parbox[t]{2mm}{\multirow{4}{*}{\rotatebox[origin=c]{90}{\small{Multimodal}}}} 
    & GIT-1vd125 & 66.5 & 60.9 & 65.6 & 52.2 & 61.3 & 51.9 & \bestall{57.8} & 48.1 & 52.6 \\ 
    ~ & Wake/Sleep & \best{73.6} & 55.6 & 64.7 & 51.4 & 61.3 & 42.0 & 50.9 & 22.8 & 38.6 \\ 
    ~ & Flamingo$_\text{CL}$ & 60.1 & 53.3 & 64.3 & 50.7 & 57.1 & 40.9 & 50.8 & 47.3 & 46.3 \\
    ~ & \emph{Best baseline: Flamingo} & 70.9 & \best{65.0} & \best{69.5} & \best{52.7} & \best{65.2} & \bestall{52.3} & 51.6 & \bestall{59.5} & \bestall{54.5} \\
\bottomrule
\end{tabular}
\caption{Macro averages for each benchmark across the top-performing systems (by overall score), best baseline, and skylines. }
\label{tab:benchmark_macro_averages}
}
\end{table*}

The results from all submissions are shown in \Cref{fig:overall-results}, with the scores of the top-performing models in each track detailed in \Cref{tab:benchmark_macro_averages}. In the figure, dashed gray lines show the performance of non-competition models (either baselines or skylines), and solid green lines show human performance on evaluation metrics. For GLUE, we use the human scores reported in \citet{nangia2019human} and for BLiMP we use the \emph{individual} human agreement scores reported in \citet{warstadt-etal-2020-blimp-benchmark-}. For Winoground, we plot the human \emph{group} score reported in \citet{Thrush_2022_CVPR}, which is slightly more stringent than our model evaluation setup as it requires humans to make the correct judgments over a set of several comparisons. For VQA, we report the \emph{Question + Image} score on \emph{real} images reported in \citet{Antol_2015_ICCV}. Again, the human task is arguably more difficult than our own evaluation as it assesses correctness in open-ended responses, rather than by comparing ground-truth captions to distractors. Therefore, the difference between the human and model scores on the vision tasks is likely an underestimate of the true difference between their respective visual capabilities.

We start our discussion by noting several high-level trends, before turning to the winning models. First, as with last year, we notice the same overall pattern of scores between our three different tracks---models in the \strict~track tend to perform better than those in the \smallstr (although the variance is higher), and models in the \vision{} track perform worse. \emph{Ceteris paribus}, more data indeed helps models learn, and learning from multimodal data remains challenging. Within text evaluations, models also perform slightly better on BLiMP compared to GLUE, which is a trend we observed last year as well.

Did model performance improve over last year? At the upper end of the distribution, the answer is \emph{yes}. This year, one model in the \smallstr track beats our Llama skyline on BLiMP, and the best model in the \strict track is within just 2.5 percentage points shy of the human score on this task. In addition to these few high-performing models, we also observed a small upward shift in the distribution of model scores compared to last year. For example, last year only 5 models in the \smallstr track achieved a GLUE score of higher than 70; this year that increased to 7 models. For the \strict track, this number was 7 last year and 8 this year. One explanation for this small upward shift is that this year we allowed contestants to bring their own data for the \strict and \smallstr tracks, provided they stayed within the data limits for each track. Many contestants modified our provided data by procuring new sources, generating data from auxiliary language models, or filtering the existing data. As we shall see in section \ref{subsec:common_methods}, dataset creation was an effective method, and we hypothesize that performance increases on our benchmark tasks over last year can be partially attributed to such data-related improvements. 

The introduction of EWoK as our hidden evaluation allowed us to observe that current systems do not learn world knowledge  within 100M words. Most submissions perform near chance, at 50\% (where dots are colored purple); the maximum score was 58.4\%.\footnote{Many masked language model submissions initially reported EWoK scores around 60--70\%. This was likely due to a default behavior of the LM evaluation harness, which assigns a label of $0$ when the probability of both sequences is the same. When changing this behavior to instead uniformly sample a label when the sequence probabilities are the same, most models get closer to 50--60\% accuracy. We confirmed these scores using a scoring script not based in the LM evaluation harness. This only affected EWoK: we were able to closely reproduce the participant-submitted scores for all other zero-shot tasks, with or without uniform sampling.} This observation highlights a potential area for future research. It may be that the current BabyLM corpus---used by many of the submitting teams---simply does not contain the world knowledge that EWoK is designed to test. One other possibility is that existing architectures have a bias towards learning linguistic phenomena more easily than relationships between concepts, physical properties, and other topics covered by EWoK. Further work on data (perhaps including data attribution methods) and algorithms will help elucidate why EWoK is so challenging for BabyLM models.

Finally, the \vision{} track proved challenging, and no submission beat the baselines we released. We discuss this further in Section \ref{sec:winning-submissions}.

\subsection{Winning Submissions} \label{sec:winning-submissions}

\paragraph{\strict{} and \smallstr{} tracks.} 

The winner of both the \strict{} and \smallstr{} tracks is GPT-BERT, submitted by \citep{charpentier_samuel_2024_bert_gpt}. GPT-BERT merges the causal (CLM) and masked language modeling (MLM) objectives from GPT and BERT, respectively, using the following key insight: by shifting MLM predictions one position to the right, the MLM predictions become aligned with next-token predictions from CLM. The authors use this insight to combine both objectives and seamlessly mix between MLM and CLM.

To train on MLM and CLM simultaneously, the authors duplicate the training data, masking and processing each copy differently for causal and masked language modeling. For each training batch, the authors choose to draw data from the CLM dataset copy with probability $p$ and from the MLM dataset with probability $1-p$. The authors explore a range of values for $p$, finding that a 1:7 causal-to-masked ratio tends to give good performance across a variety of tasks. GPT-BERT modifies the LTG-BERT architecture by adding gates on attention heads, as well as the residual connection reweighting proposed in ELC-BERT \citep{charpentier_samuel_2023_layers}, the winner of \strict and \smallstr from last year.

A different submission to this year's competition, AntLM \citep{yu_guo_2024_antlm}, also explored combining CLM and MLM by alternating between the two objectives on a per-epoch basis. The authors found that the best schedule for training LTG-BERT was 6 epochs of CLM, followed by 60 epochs of MLM, followed by 6 more epochs of CLM. While AntLM gets lower scores than GPT-BERT, it performs well overall, also beating our baselines. We conclude that 1) the LTG-BERT architecture remains a strong backbone for small language models, provided one can train it effectively, and 2) combining causal and masked language modeling objectives clearly improves performance over single objective baselines.

\paragraph{\vision{} track.} We did not award a winner for the \vision track this year. We received three submissions, and none outperformed the baselines we released. This speaks to the difficulty of multimodal learning in general. Leveraging both the text and vision modalities is challenging because the model can often learn unimodal shortcuts to solve tasks \cite{Dancette2021BeyondQB}, or the information provided by different modalities may not be aggregated properly \cite{early-vs-late-fusion2022}. Furthermore, even if there are synergistic effects from multimodal or paired inputs, such as gains in learning sample efficiency, these gains can be ephemeral given more training time \cite{zhuang-etal-2024-visual}.

While this year's \vision{} track presents what is essentially a negative result, we hope that our multimodal resources lower the barrier to entry for future research in this area. Effective methods in this space remain an unsolved challenge.

\subsection{Outstanding Paper Awards} \label{subsec:outstanding}

We presented Outstanding Paper awards to ``From Babble to Words: Pre-Training Language Models on Continuous Streams of Phonemes'' \citep{goriely_martinez_2024_babble_to_words} and ``Exploring the effect of variation sets on language model training efficiency'' \cite{haga_fukatsu_2024_variation_sets}.

We selected \citet{goriely_martinez_2024_babble_to_words} for its exploration of phonology, the study of sound or sign patterns in language, to inform tokenization. The authors incorporated phonemes into tokenization by converting raw text into phonemic transcriptions using the \texttt{phonemizer} package \cite{Bernard2021phonemizer}. They carefully ablate character-based, whitespace, and phoneme-aware tokenization schemes, ultimately arriving at a negative result: the standard BPE tokenization algorithm \citep{sennrich-etal-2016-neural} outperforms other tokenization schemes on BabyLM's text benchmarks. However, as one might expect,  phoneme-aware tokenization allows models to perform better at tasks that require phonological knowledge, such as the recognition of plausible pseudowords, or transcriptions of words that are slightly mispronounced. 

\citet{haga_fukatsu_2024_variation_sets} tackle the observation from prior work that child-directed speech improves the efficiency of training language models for certain downstream tasks, such as semantic extraction \cite{you2021child} and learning of syntactic structure \cite{mueller-linzen-2023-plant}. They hypothesize that the benefits from training on child-directed speech could be due to the existence of variation sets---consecutive rephrasings of the same sentence---which are common in child-directed speech. They construct synthetic variation sets by prompting GPT-4 for paraphrases of sentences selected from CHILDES. \citeauthor{haga_fukatsu_2024_variation_sets} find that changing the proportion of synthetic variation sets in the training data can indeed improve the performance of language models on BabyLM's evaluation tasks, although the exact characterization of this relationship remains unclear. We selected \citet{haga_fukatsu_2024_variation_sets} for the novel connections it makes between language modeling and specific theories from cognitive science.




\begin{figure}[t]
    \centering
    \includegraphics[width=0.99\linewidth]{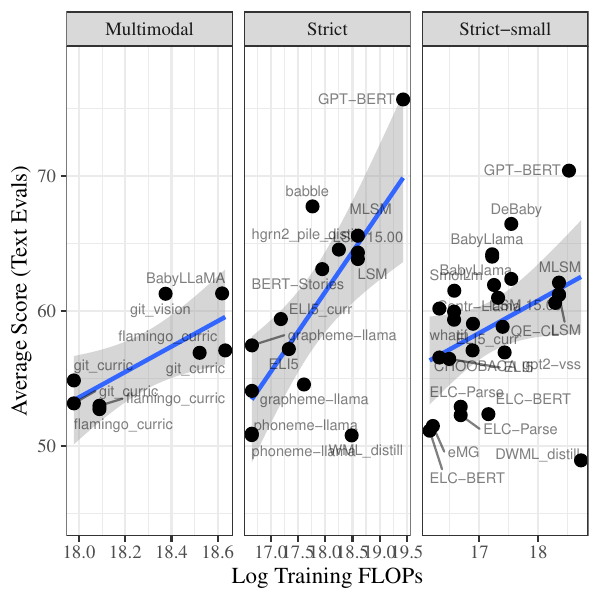}
    \caption{The relationship between training FLOPs and final score.}
    \label{fig:score_by_flops}
\end{figure}

\section{Discussion}

In this section, we discuss several trends in this year's submissions (\S \ref{subsec:flops}--\ref{subsec:common_methods}) and spotlight approaches  (\S \ref{subsec:new_approaches}) which we believe point the way towards novel and interesting work in this area. 

\subsection{Compute Budget}\label{subsec:flops}

Although we did not collect systematic metadata about last year's models, we observed that our top-performing submissions tended to be more resource-intensive, particularly in the sense that winning models were trained on a large number of epochs. This raised questions about whether their high performance was due to architectural innovations or a large compute budget. We investigate this issue further in \Cref{fig:score_by_flops}, by visualizing the relationship between models' performance on our text-only evaluations, and their total training FLOPs. We observe a positive relationship across all three tracks. To test this relationship, statistically, we fit a linear mixed-effects regression model using the \texttt{lmer4} package in \texttt{R}, with the average score on the text evaluations as our response variable, and log training FLOPs, backbone architecture and track as covariates. We included random slopes corresponding with the model's submission ID number, which indicates the research group that submitted it. We did not include interactions between the fixed effects or random slopes due to convergence issues with the model. Inspecting the fitted model, we find that more training FLOPs leads to better performance ($\beta=2.7, p<0.01$), as expected.

\begin{figure}[t]
    \centering
    \includegraphics[width=0.99\linewidth]{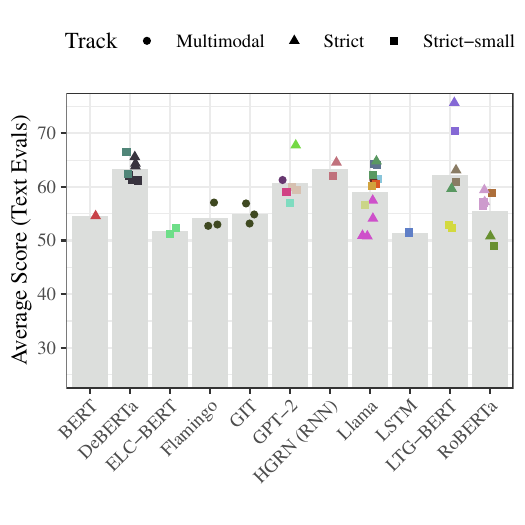}
    \caption{Scores aggregated by backbone architecture. Colors indicate different submissions.}
    \label{fig:scores_by_backbone}
\end{figure}

\subsection{Backbone Architecture} \label{subsec:arch}

In \Cref{fig:scores_by_backbone}, we visualize the averaged text evaluation score broken down by each submission's backbone architecture. Relative to last year, we received more submissions using Llama. DeBERTa and HGRN (a type of RNN) lead to the highest average scores, while the highest-scoring individual models were all based on LTG-BERT, similar to last year. To test the impact of the backbone model, we inspected the fixed effects associated with model architecture from the linear regression model described above. We found that no level of backbone architecture leads to statistically significant effects for $\alpha=0.05$, however, we did find large coefficients and smaller $p$ values for several model architectures including  DeBERTa ($\beta=9.1, p=0.06$), GPT-2 ($\beta=8.5, p=0.07$), Llama ($\beta=7.7, p=0.07$), and LTG-BERT ($\beta=8.5, p=0.06$).

Our interpretation of this result is that there are likely benefits from certain backbone architectures, but that these effects might not be strong enough to be picked up in a statistical analysis of 64 models. Interestingly, recent work has noted that different architectures and training setups often tend to converge to neural representations with similar properties and capabilities \citep{huh2024platonicrepresentationhypothesis}, and we speculate that a similar property might hold for the best models in this year's competition. 

Furthermore, different backbone architectures clearly have different variances in average text evaluation score (see Figure \ref{fig:scores_by_backbone}). This exposes another axis of architecture quality: robustness in training. For example, in this year's competition, DeBERTa \cite{he2021deberta} had high average scores, compared to other architectures, and low variance between scores in submissions. The winning architecture this year was based on LTG-BERT, but LTG-BERT also had the highest variance among all backbone architectures. This suggests that picking the ``best'' architecture might involve trading off between architectures that can achieve high scores and architectures that are straightforward to optimize and result in lower variance.


\begin{figure}[t]
    \centering
    \includegraphics[width=\linewidth]{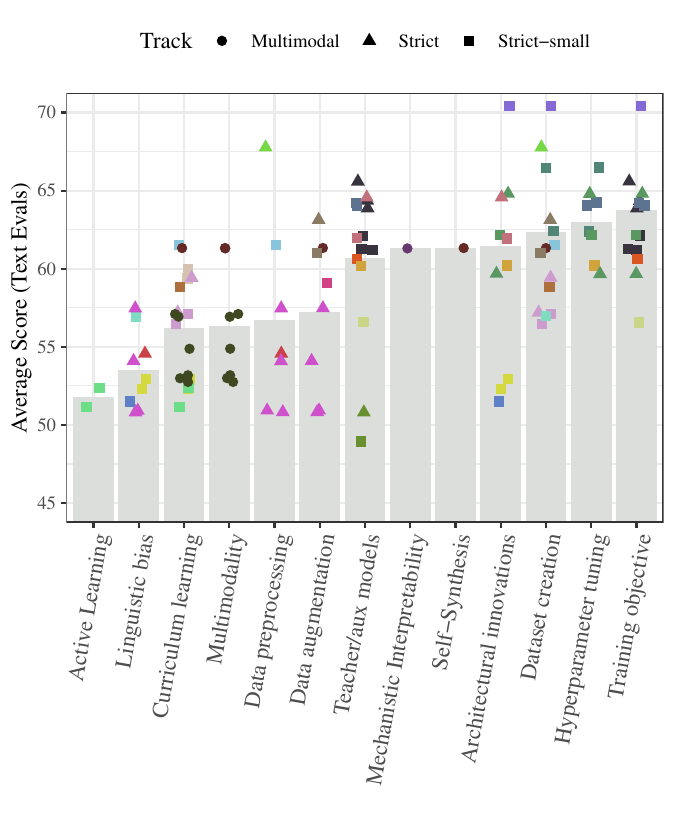}
    \vspace{-1cm}
    \caption{Scores on the BabyLM challenge, aggregated by approach. Colors indicate different submissions, which are plotted twice if they use more than one approach. Axes are zoomed to show variation in the 45-60 range more clearly.
    }
    \label{fig:scores-by_approach}
\end{figure}

\subsection{Common Methods}\label{subsec:common_methods}

In \Cref{fig:scores-by_approach} we visualize the models based on the approaches they employed. Each participant selected the categories that best fit their model, and categories were largely based on the typology of approaches we designed for analyzing the results of last year's challenge, however, we also let participants write-in approaches that we did not list.\footnote{Although some participants wrote ``controlled experiments'' and ``evaluation methods,'' we removed these from our visualization, as every team that submitted a model technically used these approaches.} Note that models are counted twice if they use more than one approach.

We find that modifications with the training objective, dataset creation, hyperparameter tuning, and architectural innovations lead to the highest average scores, although the latter also leads to a lot of variance across models. As with last year, curriculum learning, while popular, did not lead to high scores, on average. To investigate these trends more rigorously, we fit a mixed effects linear regression model in \texttt{lme4}. Our response variable was the average score for text-based evaluations, our covariates were dummy-coded variables indicating the approaches used for each model. We also included random intercepts associated with each submission ID number, corresponding to the research group that created the model. We did not include the interactions between the dummy variables due to convergence issues with the model. We found effects to be significant at $\alpha=0.05$ for four approaches: training objective innovations ($\beta=4.5, p<0.001$), dataset creation ($\beta=4.8, p<0.05$), architectural innovations ($\beta=3.5, p<0.05$), and linguistic bias ($\beta=-7.3, p<0.001$). Note that all coefficients are positive except for linguistic bias, meaning that this approach lead to \emph{lower} scores. We also found a negative effect for curriculum learning ($\beta=-3.6, p=0.055$), although the effect is not significant at the $\alpha=0.05$ level. That being said, \Cref{fig:scores-by_approach} suggests that curriculum learning is not an effective strategy for improving language models, at least in the BabyLM setting. 

\subsection{Spotlighted Approaches} \label{subsec:new_approaches}



In this section, we highlight trends and new approaches used in this year's submissions.

\paragraph{Recurrent Neural Networks (RNNs)} RNNs \cite{elman1990finding} made their debut in the BabyLM competition this year. The most effective RNN approach used the HGRN architecture \cite{qin2023hierarchically}, an RNN that adds complex forget gates on top of the Gated Recurrent Unit (GRU) architecture \cite{gru-kc}. As we noted in \S \ref{subsec:arch}, the backbone architecture, including both RNNs and Transformers, did not have a statistically significant impact on the models' performance on downstream evaluations, which is to say that the average performances across the best architectures were close. Nevertheless, RNNs and Transformers do have many differences, including their ability to express complex functions and the cost of performing inferences \cite{merrill-etal-2020-formal,merrill2024expressive}. Because RNNs may be better equipped to model human language at an algorithmic level and may be more compute effective in certain settings, it was a notable finding from this year's challenge that their performance is roughly equivalent to that of many Transformers.

\paragraph{Synthetic Data} Several contestants explored using LLMs to create synthetic training data with simple vocabularies and sentences. For example, \citet{haga_fukatsu_2024_variation_sets}, used GPT-4 to create variation sets---synthetic data that was inspired by rephrases in child-directed speech. \citet{theodoropoulos_filandrianos_2024_berttime_stories} extended the TinyStories approach \cite{eldan2023tinystoriessmalllanguagemodels}, sampling a dataset of stories using the vocabulary of a three to four-year-old child by prompting GPT-4. 

\paragraph{Corpus Construction} Since we allowed contestants to construct their own datasets, many submissions made adjustments to the baseline BabyLM corpus. Common approaches included adding data with simpler sentences and shorter words \cite{ghanizadeh_dousti_2024_data_efficient} or data better suited to certain downstream evaluations \cite{charpentier_samuel_2024_bert_gpt}. \citet{edman_bylinina_2024_babylm_second_language} viewed training corpus construction from the perspective of second language learning, skewing the training data towards sources that explain the rules of a language. 

\paragraph{Auxiliary Models} Explorations of auxiliary models and knowledge distillation were largely based on the BabyLlama approach introduced in last year's BabyLM challenge \cite{tastet_timiryasov_2024_babyllama2,yam_paek_2024_teaching_tiny_minds}. BabyLlama \cite{timiryasov-tastet-2023-baby} trains an ensemble of causal language models on a dataset and then distills the ensemble into one final model via knowledge distillation \cite{Hinton2015distillation}. Experiments revealed that BabyLlama's two-step training approach definitively outperforms simply training one causal language model \cite{tastet_timiryasov_2024_babyllama2}. \citet{berend_2024_conceptual_knowledge} used an extra training phase before pretraining, where the model learned to recover the sparsely encoded latent representation of an auxiliary model.

\paragraph{Tokenization} Along with RNNs, a new trend this year was linguistically inspired tokenization \cite{goriely_martinez_2024_babble_to_words, bunzeck_duran_2024_graphemes_vs_phonemes}. Teams explored how graphemes and phonemes could be incorporated into the language model tokenization pipeline. The primary benefit of adding graphemes and phonemes is to allow language models to perform tasks related to morphology or phonology (how words look and sound): areas where language models previously were limited \citep{babyslm}. Grapheme and phoneme-aware tokenization schemes did not seem to help language models on the base BabyLM evaluation tasks. 

\paragraph{Multi-objective training} A highly successful approach across several submissions was using multiple objectives during training. GPT-BERT and AntLM, discussed in \S \ref{sec:winning-submissions}, used different methods to combine the masked and causal language modeling objectives, and both were highly successful compared to other submissions. 

\paragraph{Training Objective Curricula} Finally, a promising variant of curriculum learning this year involved creating curricula over training objectives. \citet{salhan_martinez_2024_cross_lingual} selectively masked different parts of speech for masked language modeling over the course of training. This approach goes beyond changing the data order, which was the approach used in most curriculum learning submissions we received. We encourage participants for next year's challenge interested in curriculum learning to think beyond data order.



\section{Conclusion}\label{sec:future}

The second BabyLM Challenge has demonstrated that significant progress can be made in data-efficient language modeling through community-driven research efforts. With 31 submissions from 17 countries, the challenge revealed several key insights: innovations in model architecture, training objectives, and dataset construction proved particularly effective, with GPT-BERT, a hybrid causal-masked language model architecture, emerging as the strongest approach for the \strict and \smallstr tracks. However, the strong correlation between training FLOPs and performance suggests that computational resources remain a crucial factor even in low-data settings.

While this year's challenge added a multimodal track, in an attempt to model grounded language learning environments, no submissions outperformed the baselines in this track. This suggests that effectively integrating visual information during language learning remains a significant challenge for current architectures. This year's challenge also featured emerging research directions not present in the previous iteration, with participants exploring linguistically-motivated tokenization strategies and revisiting recurrent neural architectures. 

Looking ahead, we envision the BabyLM Challenge continuing to evolve and expand its scope beyond text-only and vision-language tracks. We hope that future iterations will explore additional modalities. such as speech, and extend to more languages, better reflecting the important fact that human language development proceeds equally well in any natural language. By broadening the challenge's focus while maintaining its core emphasis on data efficiency, we aim to inspire novel approaches that bridge the gap between artificial and human language learning. The strong participation and innovative solutions seen in this year's challenge suggest that the BabyLM community is well-positioned to tackle these ambitious goals, ultimately working toward language models that better reflect the efficiency and adaptability of human language acquisition.

\section*{Acknowledgments}

We would like to thank the organizers of CoNLL 2024 for providing us with a venue to present BabyLM. We would also like to thank the participants of the BabyLM Challenge for their innovative submissions, engagement, and contributions to the evaluation pipeline and reviewing process. 

\section*{Author Contributions}
\begin{itemize}
    \item \textbf{Primary organizers:} Alex Warstadt, Ethan Wilcox, Leshem Choshen, Aaron Mueller, Chengxu Zhuang, Michael Hu, Candace Ross
    \item \textbf{Pipeline implementation and maintenance:} Aaron Mueller
    \item \textbf{Baseline model training}: Chengxu Zhuang, Aaron Mueller
    \item \textbf{Publicity and communications with participants:} Leshem Choshen, Ethan Wilcox, Aaron Mueller, Michael Hu, Candace Ross
    \item \textbf{Training dataset compilation:} Alex Warstadt, Candace Ross, Chengxu Zhuang
    \item \textbf{Guidance on concept and workshop organization:} Ryan Cotterell, Tal Linzen, Adina Williams
    \item \textbf{Reviewing submissions:} Alex Warstadt, Ethan Wilcox, Leshem Choshen, Aaron Mueller, Michael Hu, Candace Ross
    \item \textbf{Initial draft of findings paper:} Alex Warstadt, Ethan Wilcox, Leshem Choshen, Aaron Mueller, Michael Hu, Candace Ross
    \item \textbf{Editing:} All authors
\end{itemize}

\bibliography{anthology, custom}
\bibliographystyle{acl_natbib}

\appendix
\onecolumn


\section{Text Only Datasets} \label{app:data_sources}

\paragraph{CHILDES.}
The Child Language Data Exchange System \citep[CHILDES;][]{macwhinney2000childes} is a multilingual database compiling transcriptions from numerous researchers of adult--child interactions in a range of environments, from structured laboratory activities to the home. 
\citet{huebner2021chapter} further process CHILDES, selecting only interactions with American English-speaking children ages 0--6, removing all child utterances, and tokenizing the data. 
The resulting dataset\footnote{\href{https://github.com/phueb/BabyBERTa/blob/master/data/corpora/aochildes.txt}{\url{https://github.com/phueb/BabyBERTa/blob/master/data/corpora/aochildes.txt}}} contains about 5M words.

\paragraph{British National Corpus.}
The BNC \citep{bncconsortium2007british} is a 100M word multi-domain corpus of British English from the second half of the 20$\textsuperscript{th}$ century. 
We select only the dialogue portion of the corpus, totaling about 10M words.

\paragraph{Children's Book Test.}
CBT is a compilation of over a hundred children's books from Project Gutenberg by \citet{hill-2016-cbt}. 
The dataset was originally released with a set of questions for testing named entity prediction, which we do not include in the pretraining data.

\paragraph{Children's Stories Text Corpus.}
This dataset consists of manually selected children's stories from Project Gutenberg. 
It was compiled by \citet{bensaid2021fairytailor} for the development of a story generation system. 

\paragraph{Project Gutenberg.}
The Standardized Project Gutenberg Corpus \citep{gerlach2020standardized} is a curated and preprocessed selection of over 50k literary books in the public domain from Project Gutenberg totaling over 3B tokens.%
\footnote{\href{https://gutenberg.org/}{\url{https://gutenberg.org/}}}
This distribution comes with extensive metadata that allows us to filter texts by language and date.

\paragraph{OpenSubtitles.}
This dataset \citep{lison2016opensubtitles} is a compilation of publicly available subtitles from TV and movies on a third-party website.%
\footnote{\href{http://opensubtitles.org/}{\url{http://opensubtitles.org/}}} 
We use only the English portion.

\paragraph{Wikipedia.}
Wikipedia is a volunteer-authored encyclopedia hosted by the Wikimedia Foundation.
We use only the English portion.

\paragraph{Simple English Wikipedia.}
Simple English is classified as a separate language in Wikipedia, thus the texts here are disjoint from those in English Wikipedia.
The texts use shorter sentences and high-frequency vocabulary and avoid idioms.

\paragraph{Switchboard Corpus.}
The Switchboard Corpus \citep{godfrey1992switchboard} is a collection of transcribed telephone conversations between pairs of strangers.
We accessed the text through the Switchboard Dialog Act Corpus \citep{Stolcke-etal:2000}.

\subsection{Text--Image Datasets}
The corpus for the \vision{} track consisted of 50M words from the above datasets, as well as 50M more from image-caption datasets. These include the following:

\paragraph{Localized Narratives.}
Localized Narratives \citep{LocalizedNarratives} is an image-caption dataset. Images are labeled by human annotators; the annotators were asked to describe an image with their voice while hovering their mouse over the region being described. We use the MS-COCO and Open Images subsets.

\paragraph{Conceptual Captions.}
Conceptual Captions \citep{sharma2018conceptual} is an image-capture dataset consisting of automatically scraped and filtered images and captions/annotations from billions of web pages.

\begin{minipage}{\textwidth}
\setlength{\parindent}{10mm}

\section{Evaluation Data Details}\label{app:eval-data-size}

As described in \Cref{sec:eval-pipeline}, we filtered out evaluation examples containing words that did not appear at least twice in both the \smallstr{} and \vision{} pretraining corpora. Here, we present the number of training and test examples for each evaluation task after filtering.

Note that we only control for lexical content: other factors, such as sentence length, syntactic complexity, and overall linguistic style, remain distinct between our corpus and these tasks. In the future, it would be helpful for researchers to focus on designing tasks on which both children \emph{and} language models can be reasonably evaluated.

Note, too, that this filtering step implies that we cannot directly compare results obtained from the BabyLM Challenge to prior evaluations using the full datasets. We also cannot directly compare to results from last year's challenge, though we believe the overlap between the evaluation sets across the BabyLM Challenges is likely high.

\vspace{3em}

    \centering
    \resizebox{0.48\linewidth}{!}{
    \begin{tabular}{clrr}
    \toprule
        \textbf{Task} & \textbf{Subtask} & \textbf{$\mid$Train$\mid$} & \textbf{$\mid$Test$\mid$} \\
    \midrule
        BLiMP & -- & -- & 59875 \\
    \midrule
        \parbox[t]{2mm}{\multirow{5}{*}{\rotatebox[origin=c]{90}{\small{BLiMP}}}} \parbox[t]{2mm}{\multirow{5}{*}{\rotatebox[origin=c]{90}{\small{Supplement}}}} & Hypernym & -- & 842 \\
        & Question-Answer Congruence (easy) & -- & 64 \\
        & Question-Answer Congruence (tricky) & -- & 165 \\
        & Subject-Auxiliary Inversion & -- & 3867 \\
        & Turn-taking & -- & 280 \\
    \midrule
        \parbox[t]{2mm}{\multirow{11}{*}{\rotatebox[origin=c]{90}{\small{(Super)GLUE}}}} & CoLA & 8551 & 522 \\
        & SST-2 & 67349 & 436 \\
        & MRPC & 3668 & 204 \\
        & QQP & 363846 & 20215 \\
        & MNLI & 392702 & 4908 \\
        & MNLI-mismatched & -- & 4916 \\
        & QNLI & 104743 & 2732 \\
        & RTE & 2490 & 139 \\
        & BoolQ & 9427 & 1635 \\
        & MultiRC & 27243 & 2424 \\
        & WSC & 554 & 52 \\
    \midrule
        \parbox[t]{2mm}{\multirow{11}{*}{\rotatebox[origin=c]{90}{\small{EWoK}}}} & Agent Properties & -- & 2210 \\
        & Material Dynamics & -- & 770 \\
        & Material Properties & -- & 170 \\
        & Physical Dynamics & -- & 120 \\
        & Physical Interactions & -- & 556 \\
        & Physical Relations & -- & 818 \\
        & Quantitative Properties & -- & 314 \\
        & Social Interactions & -- & 294 \\
        & Social Properties & -- & 328 \\
        & Social Relations & -- & 1548 \\
        & Spatial Relations & -- & 490 \\
    \bottomrule
    \end{tabular}}
    \resizebox{0.48\linewidth}{!}{
    \begin{tabular}{clrr}
    \toprule
        \textbf{Task} & \textbf{Subtask} & \textbf{$\mid$Train$\mid$} & \textbf{$\mid$Test$\mid$} \\
    \midrule
        VQA & -- & -- & 25230 \\
    \midrule
        Winoground & -- & -- & 746 \\
    \midrule
        \parbox[t]{2mm}{\multirow{3}{*}{\rotatebox[origin=c]{90}{\small{DevBench}}}} & Visual Vocabulary & -- & 433 \\
        & Test of Receptive Grammar (TROG) & -- & 79 \\
        & THINGS & -- & 12340 \\
    \bottomrule
    \end{tabular}}
    \captionof{table}{Number of training and test examples for each BabyLM evaluation task. We present the number of examples for the text-only tasks (left) and the multimodal tasks (right). We show the number of examples \emph{after} filtering based on the pre-training corpus vocabulary (\Cref{sec:eval-pipeline}). Note that only the (Super)GLUE has training examples; the rest of the tasks are zero-shot.}
    \label{tab:eval-data-size}

\end{minipage}

\section{Subtask Results}
Here, we present a more detailed breakdown of results by subtask. Each task has a subsection containing a table where results are described, as well as a textual description containing and overview of the main takeaways for each task.

\subsection{BLiMP and BLiMP Supplement}
\setlength{\parindent}{5mm}
GPT-BERT was the best-performing model on the BLiMP tasks in both the \strict{} and \smallstr{} tracks. The only subtask where it did not perform best among all models was for Hypernym, where the LTG-BERT baseline was best. BabbleGPT and AntLM were the runners-up in the \strict{} track, whereas DeBaby and BabyLlama-2 were the runners-up in the \smallstr{} track. In general, submissions to the \vision{} track did not consistently outperform the baseline models; Wake/Sleep outperformed the best baseline (Flamingo) on BLiMP, but no submission outperformed Flamingo on the BLiMP Supplement.

In general, the average BLiMP score across subtasks was effective in distinguishing between high- and low-performing systems: there is high variance across submissions, and those that perform best on BLiMP also tend to perform comparatively well on other tasks.

Similarly to last year, we observe that the \testsuite{hypernym} test suite is beyond the ability of language models of this scale.
All models (including last year's skylines) perform very close to chance, suggesting either that their preferences are virtually random guessing, or they show systematic biases that essentially cancel out due to counterbalancing in the test data.
However, we hesitate to conclude that these models have no knowledge of lexical entailment relations for two reasons:
First, these test sentences are somewhat unnatural logical statements that are out-of-domain for the models; and
second, there is less reason \emph{a priori} to believe that logically invalid statements have lower probabilities than valid statements.

Among the \testsuite{question--answer congruence} test suites, we find that the ``tricky'' set is still highly discriminative, probably due in part to its adversarial nature. This tells us that most models are easily fooled by locally coherent distractor answers and pay too little attention to cross-sentential long-distance dependency between a \emph{wh}-word and a congruent answer.
Only the top-performing models in the \strict{} track score better than chance, and the RoBERTa skyline outperforms all models by a wide margin.

The tests for \testsuite{subject--auxiliary inversion} are relatively easy: the best models reach near-perfect accuracy, and all models score relatively high compared to other test suites.

Finally, \testsuite{turn taking} is highly discriminative, with some models performing at or near chance, while the best model achieves accuracy over 90\%.

\def\supptask#1{\rotatebox[origin=c]{90}{\parbox[t]{6em}{\small#1}}}
{\centering
\begin{tabular}{lllllllll}
\toprule
    & & BLiMP & \multicolumn{6}{c}{BLiMP Supplement} \\\cmidrule(lr){3-3}\cmidrule(lr){4-9}
    \textbf{} & \textbf{Model} & 
    \supptask{Macro\\average} & \supptask{Macro\\average} &
    \supptask{Hypernym} & 
    \supptask{Q--A congruence\\(easy)} & 
    \supptask{Q--A congruence\\(tricky)} & 
    \supptask{Subject--aux\\inversion} & 
    \supptask{Turn taking} \\ 
\midrule
    \parbox[t]{2mm}{\multirow{4}{*}{\rotatebox[origin=c]{90}{\small{Strict}}}} 
    & GPT-BERT & \bestall{86.1} & \bestall{76.8} & 48.8 & \bestall{90.6} & \bestall{59.4} & \bestall{96.3} & \bestall{88.9} \\ 
    ~ & BabbleGPT & 77.8 & 69.5 & 47.9 & 81.2 & 52.1 & 81.9 & 84.3 \\ 
    ~ & AntLM & 74.9 & 66.0 & 49.3 & 79.7 & 43.6 & 78.3 & 79.3 \\ 
    ~ & \emph{Base baseline: LTG-BERT} & 69.2 & 66.5 & \bestall{55.0} & 75.0 & 53.3 & 87.5 & 61.4 \\ 
\midrule
    \parbox[t]{2mm}{\multirow{4}{*}{\rotatebox[origin=c]{90}{\small{Strict-small}}}} 
    & GPT-BERT & \best{81.2} & \best{69.4} & 47.1 & 73.4 & \best{54.5} & 86.3 & \best{85.7} \\ 
    ~ & DeBaby & 74.2 & 63.7 & \best{53.3} & \best{79.7} & 49.1 & 84.1 & 52.1 \\ 
    ~ & BabyLlama-2 & 73.2 & 63.1 & 49.8 & 59.4 & 41.2 & \best{90.3} & 75.0 \\ 
    ~ & \emph{Best baseline: BabyLlama} & 69.8 & 59.5 & 49.6 & 54.7 & 41.2 & 86.0 & 66.1 \\ 
\midrule
    \parbox[t]{2mm}{\multirow{4}{*}{\rotatebox[origin=c]{90}{\small{Multimodal}}}} 
    & Wake/Sleep & \best{73.6} & 55.6 & \best{49.5} & 50.0 & 30.9 & 85.3 & 62.1 \\ 
    ~ & GIT-1vd125 & 66.5 & 60.9 & 48.2 & 57.8 & \best{44.2} & \best{86.5} & 67.9 \\ 
    ~ & GIT$_\text{CL}$ & 64.0 & 51.2 & 48.9 & 50.0 & 20.0 & 83.7 & 53.6 \\ 
    ~ & \emph{Best baseline: Flamingo} & 70.9 & \best{65.0} & 48.8 & \best{75.0} & 43.6 & 86.2 & \best{71.4} \\
\bottomrule
\end{tabular}
\captionof{table}{BLiMP Supplement accuracies for each subtask for the top performing systems (by overall score), best baseline, and skylines. For each subtask, we mark the \best{best} performing system for each track, and the  \bestall{best} performing system overall.}
\label{tab:blimp_subtasks}
}

\subsection{GLUE/SuperGLUE}
Scores on (Super)GLUE tasks (Table~\ref{tab:glue_subtasks}) show that GPT-BERT is the best-performing system in both the \strict and \smallstr{} tracks. Notably, its performance in the \smallstr{} track is better than the runners-up in the \strict{} track, suggesting that this approach is highly data-efficient and/or well-tuned for small-scale language modeling. BabbleGPT and AntLM were again the runners-up for (Super)GLUE in the \strict{} track, and DeBaby was again the runner-up for the \smallstr{} track. MLSM is now second runner-up in the \smallstr{} track. Once again, no submissions outperformed the best baseline (Flamingo) in the \vision{} track. This largely confirms findings from the BLiMP and BLiMP Supplement tasks.

\def\gluetask#1{\rotatebox[origin=c]{90}{\parbox[t]{3.5em}{\small\textbf{#1}}}}
\centering
\resizebox{\textwidth}{!}{%
\begin{tabular}{llllllllllllll}
\toprule
    ~ & \textbf{Model} & 
    \gluetask{Macro\\average} & 
    \gluetask{CoLA} & 
    \gluetask{SST-2} & 
    \gluetask{MRPC} & 
    \gluetask{QQP} & 
    \gluetask{MNLI} & 
    \gluetask{MNLI-mm} & 
    \gluetask{QNLI} & 
    \gluetask{RTE} & 
    \gluetask{BoolQ} & 
    \gluetask{MultiRC} & 
    \gluetask{WSC} \\ 
\midrule
    \parbox[t]{2mm}{\multirow{4}{*}{\rotatebox[origin=c]{90}{\small{Strict}}}} & GPT-BERT & \bestall{81.5} & \bestall{62.4} & \bestall{94.0} & \bestall{94.4} & \bestall{89.1} & \bestall{85.2} & \bestall{85.3} & \bestall{90.8} & \bestall{69.1} & \bestall{78.4} & \bestall{73.3} & \bestall{75.0} \\ 
    ~ & Babble-GPT & 71.7 & 37.8 & 89.4 & 83.8 & 84.0 & 75.3 & 76.4 & 82.9 & 66.2 & 63.7 & 65.1 & 63.5 \\ 
    ~ & AntLM & 66.3 & 22.2 & 89.4 & 84.9 & 84.2 & 74.8 & 74.4 & 83.2 & 55.4 & 65.8 & 59.9 & 34.6 \\ 
    ~ & \emph{Best baseline: LTG-BERT} & 68.4 & 34.6 & 91.5 & 83.1 & 86.7 & 77.7 & 78.1 & 78.2 & 46.8 & 61.7 & 52.6 & 61.5 \\ 
\midrule
    \parbox[t]{2mm}{\multirow{4}{*}{\rotatebox[origin=c]{90}{\small{Strict-small}}}} & GPT-BERT & \best{76.5} & \best{48.9} & \best{92.2} & \best{91.5} & \best{87.1} & \best{80.2} & \best{80.5} & \best{86.4} & 64.0 & \best{72.5} & \best{69.3} & \best{69.2} \\ 
    ~ & DeBaby & 73.7 & 41.8 & 89.2 & 91.2 & 86.6 & 78.1 & 77.6 & 85.5 & \best{69.8} & 71.1 & 64.2 & 55.8 \\ 
    ~ & MLSM & 73.3 & 45.2 & 90.6 & 82.2 & 86.6 & 76.4 & 77.4 & 84.7 & 60.4 & 69.4 & 67.6 & 65.4 \\
    ~ & \emph{Best baseline: BabyLlama} & 63.3 & 2.2 & 86.2 & 82.0 & 83.6 & 72.4 & 74.2 & 82.8 & 49.6 & 65.0 & 60.1 & 38.5 \\ 
\midrule
    \parbox[t]{2mm}{\multirow{4}{*}{\rotatebox[origin=c]{90}{\small{Multimodal}}}} & GIT-1vd125 & 65.6 & 30.7 & 89.7 & 81.5 & 83.3 & 72.7 & 72.6 & 78.4 & 51.8 & 64.2 & 54.7 & 42.3 \\ 
    ~ & Wake/Sleep & 64.7 & 12.2 & 79.8 & 78.4 & 80.5 & 69.4 & 70.6 & 79.8 & 52.5 & 63.1 & \best{65.8} & \best{59.6} \\ 
    ~ & Flamingo$_\text{CL}$ & 64.3 & 31.8 & 88.3 & 82.4 & 81.9 & 70.4 & 71.4 & 69.9 & 46.0 & 66.5 & 56.2 & 42.3 \\
    ~ & \emph{Best baseline: Flamingo} & \best{69.5} & \best{36.7} & \best{90.4} & \best{84.2} & \best{85.1} & \best{75.8} & \best{76.4} & \best{83.8} & \best{60.4} & \best{69.1} & 60.5 & 42.3 \\
\bottomrule
\end{tabular}}
\captionof{table}{(Super)GLUE results for each subtask for the top performing systems (by overall score), best baseline, and skylines. For each subtask, we mark the \best{best} performing system for each track, and the \bestall{best} performing system overall.}
\label{tab:glue_subtasks}

\subsection{Multimodal Tasks}

\def\subtask#1{\rotatebox[origin=c]{90}{\parbox[t]{5em}{\small\textbf{#1}}}}
\begin{tabular}{llll}
\toprule
    \textbf{Model} & 
    \subtask{Macro\\average} & 
    \subtask{VQA} & 
    \subtask{Winoground} \\
\midrule
    GIT-1vd125 & \bestall{54.9} & 51.9 & \bestall{57.8} \\ 
    GIT$_\text{CL}$ & 49.6 & 44.0 & 55.2 \\ 
    Wake/Sleep & 46.5 & 42.0 & 50.0 \\ 
    \emph{Best baseline: GIT} & 54.8 & \bestall{54.1} & 55.5 \\ 
\bottomrule
\end{tabular} 
\captionof{table}{Results for the public multimodal tasks for the top performing systems (by average score), and the best baseline. For each subtask, we mark the \best{best} performing system for each track, and the \bestall{best} performing system overall.}

\def\subtask#1{\rotatebox[origin=c]{90}{\parbox[t]{4em}{\small\textbf{#1}}}}
\begin{tabular}{lllll}
\toprule
    \textbf{Model} & 
    \subtask{Macro\\average} & 
    \subtask{Visual\\Vocabulary} & 
    \subtask{TROG} & 
    \subtask{THINGS} \\
\midrule
    Flamingo$_\text{CC}$ & 49.0 & 66.4 & 34.2 & \bestall{46.5} \\ 
    GIT$_\text{CL}$ & 48.2 & 73.1 & \bestall{39.5} & 32.1 \\
    GIT-1vd125 & 48.1 & \bestall{84.9} & 35.5 & 23.8 \\ 
    \emph{Best baseline: Flamingo} & \bestall{59.5} & 80.7 & 38.2 & 32.6 \\ 
\bottomrule
\end{tabular} 
\captionof{table}{Results for the DevBench tasks for the top performing systems (by average score), and the best baseline. For each subtask, we mark the \best{best} performing system for each track, and the \bestall{best} performing system overall.}
\label{tab:vision_subtasks}

\end{document}